\documentclass[letterpaper, 10pt, conference]{ieeeconf}

\IEEEoverridecommandlockouts
\usepackage{graphicx}
\usepackage{subcaption}

\usepackage{times}

\usepackage{amsmath}
\usepackage{amssymb}

\usepackage{booktabs}
\usepackage{multirow}
\usepackage{tabularx}

\usepackage{algorithm}
\usepackage{algpseudocode}

\usepackage{xcolor}
\usepackage{float}
\usepackage{romannum}

\usepackage{cite}
\usepackage{url}
\usepackage{hyperref}

\title{\LARGE \bf
CA-AC-MPC: CUDA-Accelerated Actor-Critic Model Predictive Control}

\author{Antonio Buo, Vittorio Cammarota, Michele Avagnale, Pierluigi Arpenti, Vincenzo Lippiello, Fabio Ruggiero
\thanks{This work was partially supported by the Italian Ministry of University and Research, under the complementary actions to the NRRP 'Fit4MedRob - Fit for Medical Robotics' Grant (\# PNC0000007). This research was also partially supported by the INVITALIA Project NEMESI (CUP C67G22000420008). }
\thanks{$^{1}$Authors are with PRISMA Lab and CREATE Consortium, Department of Electrical Engineering and Information Technology, University of Naples Federico II, Naples, Italy. Corresponding author's email: {\tt\small [fabio.ruggiero@unina.it]}.}
}
\begin{document}
\maketitle
\thispagestyle{empty}
\pagestyle{empty}

\begin{abstract}
In the literature, actor-critic model predictive control  (AC-MPC) integrates MPC with reinforcement learning to enable high-performance control of complex dynamical systems. However, its differentiable MPC layer requires repeatedly solving an optimization problem in both the forward and backward passes, leading to substantial training and inference latency. This paper tackles this bottleneck introducing a CUDA-accelerated variant that significantly reduces end-to-end execution time while preserving the control performance of the baseline formulation. Simulation results on an agile drone racing task show that our approach achieves state-of-the-art lap times and near-limit dynamic behaviour with markedly reduced training and inference time.
\end{abstract}

\section*{Supplementary Material}
Code can be found at https://github.com/prisma-lab/CA-AC-MPC.
\section*{List of most used acronyms}
\noindent
\small
\begin{tabularx}{\columnwidth}{@{}p{0.26\columnwidth}X@{}}
\hline
\textbf{Acronym} & \textbf{Meaning} \\
\hline
AC-MPC & Actor--Critic Model Predictive Control \\
AC-MLP & Actor--Critic Multilayer Perceptron \\
CA-AC-MPC & CUDA-Accelerated Actor--Critic Model Predictive Control \\
CA-DiffMPC & CUDA-Accelerated Differentiable Model Predictive Control \\
CUDA & Compute Unified Device Architecture \\
DiffMPC & Differentiable Model Predictive Control \\
MLP & Multilayer Perceptron \\
MPC & Model Predictive Control \\
\hline
\end{tabularx}
\normalsize
\section{Introduction}\label{sec:intro}
Control design is especially difficult in high-performance robotic applications, where robots operate in dynamic conditions with unmodelled effects, parameter uncertainties, and rapid transients.
The bottleneck of classical model-based control strategies---including online optimization methods such as MPC---is not the model-based paradigm per se, but the mismatch between model fidelity and computational tractability~\cite{xie2024three}. Pushing performance often requires dynamical models that are either too inaccurate to capture the relevant phenomena or too expensive to be used in real-time~\cite{romero2022tro}.

This gap motivated growing interest in reinforcement learning (RL) and other data-driven approaches, which can optimize control policies directly from task performance without requiring an explicit high-fidelity model.
While RL still requires significant design choices (e.g., reward shaping, curriculum design, hyperparameter tuning, etc.), it provides a flexible framework in which feedback policies are learned through interaction with the environment. In particular, RL can handle non-differentiable and structured objectives without relying on proxy references such as predefined trajectories or paths. This has enabled state-of-the-art results in highly dynamic robotic tasks where model-free RL has been shown to outperform classical optimal control in near-limit conditions~\cite{song2023reaching,kaufmann2023champion}. However, RL approaches typically require massive amounts of data---often millions of simulated interactions.

These challenges have led research to focus on hybrid control frameworks that combine the structure and reliability of model-based optimal control with the flexibility and task-driven optimization of RL. Among these, AC-MPC emerged as a promising paradigm, in which an MPC-based actor provides short-horizon optimal control actions while a learned critic captures long-term value information~\cite{romero2024icra}. This integration yields policies that retain the interpretability and constraint-handling capabilities of MPC while benefiting from the adaptability and performance of RL~\cite{romero2022tro}.
Nevertheless, embedding an optimization-based control layer within a neural policy introduces a computational burden, since each policy evaluation requires solving a parameterized optimal control problem (OCP) rather than executing a simple feed-forward pass~\cite{amos2017optnet,amos2018differentiablempc}. Moreover, MPC typically assumes access to an accurate \emph{differentiable} dynamic model~\cite{amos2018differentiablempc,romero2022tro}. 
In the AC-MPC, the OCP must be solved during the forward pass to compute the control action, and an additional linear system must be solved during the backward pass to propagate gradients, increasing training time. During inference, gradients are not required and only the forward MPC pass is performed; however, the policy remains substantially slower than a standard AC-MLP~\cite{romero2024icra}. 


This paper addresses this limitation by introducing a CUDA-accelerated actor-critic model predictive control framework, referred to as CA-AC-MPC.
Unlike the original PyTorch-based implementation \cite{amos_mpc_pytorch,romero2022tro}, which relies on PyTorch’s GPU backend, our approach directly invokes CUDA kernels for the most computationally intensive operations. This provides finer-grained control over GPU parallelism, reduces overhead, and accelerates both the forward and backward passes of the differentiable MPC layer.

\begin{figure}[t!]
\centering
\includegraphics[width=\columnwidth]{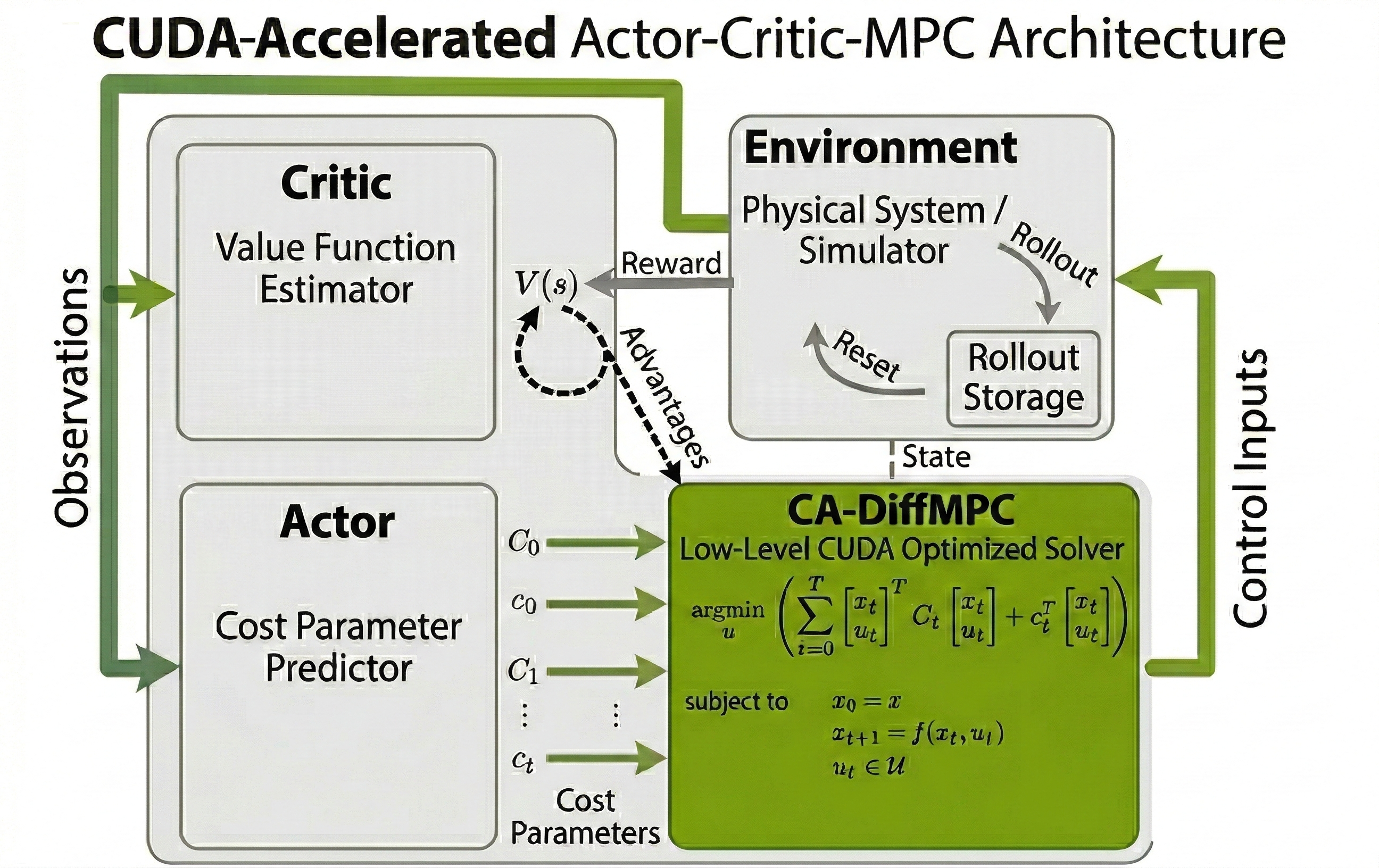}
\caption{Inspired from the initial picture in~\cite{romero2024icra}, we present the overview of the proposed CA-AC-MPC framework, highlighting the introduced low-level CUDA optimized solver.}
\label{fig:CA-diffMPPC}
\vspace{-0.5cm}
\end{figure}

The specific contribution of this work is a fused-kernel CUDA implementation of iterative linear quadratic regulator (iLQR) for differentiable MPC. We present a C++/CUDA PyTorch extension that restructures the iLQR computation into three fused stages (rollout and linearization, Riccati backward pass with box constraints, and line-search rollout). This replaces the Python-level recursion over the horizon with a fixed number of CUDA kernel launches per iteration. We refer to this implementation as CA-DiffMPC. When integrated into the AC-MPC framework, CA-DiffMPC gives rise to our CA-AC-MPC (see Fig.~\ref{fig:CA-diffMPPC}), yielding an overall speed-up of one to two orders of magnitude in our simulations, while also reducing runtime inference.
The proposed framework is validated on an agile drone-racing task, where CA-AC-MPC achieves lap-time performance comparable to, and in some cases better than, an end-to-end AC-MLP baseline, consistent with the performance reported for AC-MPC in~\cite{Romero_2026}.
We release the source code to facilitate future research\footnote{https://github.com/prisma-lab/CA-AC-MPC.}.
\section{Related Works}\label{sec:related_works}
The integration of MPC with RL has its roots in differential dynamic programming and dual control. 
Early approaches such as guided policy search~\cite{levine2016gps} and MPC-guided policy learning~\cite{nagabandi2018mpc} leveraged MPC as a supervisory signal or data generator for policy training. Recent developments in DiffMPC~\cite{amos2018differentiablempc} have enabled the inclusion of OCPs as differentiable layers within neural networks, allowing gradient-based policy optimization~\cite{romero2024icra}.

Nevertheless, DiffMPC-based approaches remain computationally demanding. In particular, end-to-end training of the actor requires computing gradients of the MPC solution with respect to the cost parameters predicted by the neural network. Since the MPC solver does not define a closed-form mapping, naive automatic differentiation through the solver iterations (i.e., unrolling) is both computationally expensive and numerically unstable~\cite{agrawal2019differentiable}. An alternative is implicit differentiation, which computes gradients by differentiating the Karush–Kuhn–Tucker (KKT) optimality conditions at convergence, thus avoiding the need to store the full computational graph~\cite{amos2018differentiablempc}. 
A complementary direction to mitigate this computational burden is to exploit parallel hardware acceleration. 
For instance, a GPU-based MPC framework for legged locomotion that enables both temporal and state-space parallelization via a parallel associative scan to solve the primal–dual KKT system is proposed in~\cite{amatucciral2026}.
Recently, a GPU-accelerated DiffMPC solver based on sequential quadratic programming, combined with a custom preconditioned conjugate gradient routine and structure-exploiting tridiagonal preconditioning, was introduced in~\cite{adabag2025differentiable}, enabling efficient parallelization of the backward pass.
However, existing approaches predominantly address the computational bottleneck of DiffMPC at the algorithmic level~\cite{adabag2025differentiable,amatucciral2026}. 
As acknowledged in those works, GPU acceleration can become ineffective---and in some cases slower than CPU-based implementations---for small- to medium-scale problems, where temporal parallelism does not amortize kernel launch overhead because the per-time step arithmetic is relatively cheap. This regime is highly relevant in robotics, where state and control dimensions are often moderate and dispatch latency can dominate end-to-end execution time. 
The proposed CA-AC-MPC targets this dispatch overhead directly. 
\section{Methodological Background}\label{sec:method_back}
In this section, the fundamental concepts underlying the AC-MPC framework are briefly summarized. The reader can refer to the cited works for a more detailed discussion.
\subsection{Economic Model Predictive Control}
Let \(n_x \in \mathbb{N}\) and \(n_u \in \mathbb{N}\) denote the numbers of states and control inputs, respectively, of a given discrete-time dynamical system with nonlinear state-transition map \(f(\cdot,\cdot): \mathbb{R}^{n_x} \times \mathbb{R}^{n_u} \to \mathbb{R}^{n_x}\), i.e.,
$x_{t+1} = f(x_t, u_t)$, where $x_t \in \mathcal{X} \subseteq \mathbb{R}^{n_x}$ denotes the state of the system, belonging to the set $\mathcal{X}$, at time $t$, and $u_t \in \mathcal{U} \subseteq \mathbb{R}^{n_u}$ denotes the control input of the system, belonging to the set $\mathcal{U}$, at time $t$. 
Consider the discrete-time finite-horizon OCP~\cite{borrelli2017predictive}
\begin{equation}\label{eq:FHOCP}
\begin{aligned}
\underset{u_0,\dots,u_{T-1}}{\operatorname{min}}\;&
\sum_{t=0}^{T-1}\ell_t(x_t,u_t)+V_f(x_T)\\
\text{s.t.}\quad
& x_{t+1}=f(x_t,u_t),\quad t=0,\dots,T-1\\
&u_t\in\mathcal{U},\quad t=0,\dots,T-1\\
&x_0=x_{\mathrm{init}}
\end{aligned}
\end{equation}
where $\ell(\cdot,\cdot) : \mathbb{R}^{n_x} \times \mathbb{R}^{n_u} \to \mathbb{R}$ is the stage cost, $V_f(\cdot): \mathbb{R}^{n_x} \to \mathbb{R}$ is the terminal cost, and $x_{\mathrm{init}} \in \mathbb{R}^{n_x}$ and $x_{T} \in \mathbb{R}^{n_x}$ are the initial and terminal states of the system, respectively.
In MPC, at each sampling instant, the finite-horizon optimization problem~\eqref{eq:FHOCP} is solved using the current state at time step \(t\) as the initial condition \(x_{\mathrm{init}}\)~\cite{grne2013nonlinear}. Subsequently, only the first control input \(u_0^\ast\) of the resulting optimal control sequence is applied to the closed-loop system. The procedure is then repeated at the next time step using the newly measured state\cite{borrelli2017predictive}. 

In standard tracking MPC, the stage cost \(\ell_t(\cdot,\cdot)\) is typically chosen as a positive-definite quadratic function that penalizes deviations from a prescribed trajectory/setpoint. A broader class of formulations is referred to as economic MPC~\cite{ellis2017empc}, in which the tracking cost is replaced by a more general stage cost encoding a task-specific performance objective. This flexibility introduces significant challenges. In many tasks, the stage cost \(\ell_t(\cdot,\cdot)\) is sparse, non-smooth, or non-differentiable, complicating both the formulation and the numerical solution of the OCP. Indeed, performance indicators of practical interest are often not smooth by nature~\cite{Findeisen2017EMPC}: they may exhibit discontinuities, piecewise definitions, or abrupt changes depending on the operating regime. 
Hence, constructing a globally smooth and informative cost function is often difficult.
The lack of differentiability is particularly problematic for gradient-based optimization method. This often requires smoothing approximations, differentiable surrogate costs, or dedicated non-smooth optimization techniques.
A common strategy to address these issues in economic MPC is to approximate the original stage cost with a quadratic surrogate that is more amenable to numerical optimization. Defining $ z_t=\begin{bmatrix}x_t^\top & u_t^\top\end{bmatrix}^\top \in \mathbb{R}^{n_x+n_u}$,
the stage cost can be approximated as
\begin{equation}\label{eq:stage-cost}
\ell(z_t)\approx \tfrac{1}{2} z_t^\top C_t z_t + c_t^\top z_t,
\end{equation}
where \(C_t \in \mathbb{R}^{(n_x+n_u)\times(n_x+n_u)}\) is a positive semidefinite cost matrix and \(c_t\in\mathbb{R}^{n_x+n_u}\) collects the cost parameters. 
The quadratic surrogate in~\eqref{eq:stage-cost} yields a smooth objective and makes the OCP differentiable with respect to \((C_t,c_t)\), which is essential when embedding MPC as a differentiable layer within a learning pipeline. 
However, in economic control tasks, the relevant objective is often time-varying and context-dependent~\cite{ellis2017empc}; therefore, fixed hand-tuned parameters may fail to generalize across operating regimes.

\subsection{Actor–Critic Policy Architecture}
In RL, the interaction between an agent and its environment is commonly formalized as a Markov decision process (MDP), described by the tuple $\mathcal{M}=\langle \mathcal{S},\,\mathcal{A},\,P,\,R,\,\gamma \rangle$, where \(\mathcal{S}\) is the state space (with states \(s\in\mathcal{S}\)), \(\mathcal{A}\) is the action space (with actions \(a\in\mathcal{A}\)), \(P(s'\mid s,a)=\Pr(s_{t+1}=s' \mid s_t=s, a_t=a)\) is the state-transition probability, \(R(s,a)=\mathbb{E}[r_t \mid s_t=s, a_t=a]\) is the expected immediate reward function (with reward $r$), and \(\gamma\in[0,1]\) is the discount factor.
One way to solve these MDP problems is by using temporal difference (TD) methods. The class of TD algorithms updates a value estimate $V(s)$ toward a target that combines the observed reward and the estimate at the next state. The update for a state-value function under policy $\pi$ is given by $V(s_t) \leftarrow V(s_t) + \alpha \left[r_t + \gamma V(s_{t+1}) - V(s_t)\right]$where $\alpha>0 \in \mathbb{R}$ is a learning rate and the temporal-difference error $\delta_t = r_t + \gamma V(s_{t+1}) - V(s_t)$ quantifies the mismatch between successive value estimates\cite{712192}.
The actor–critic class of algorithms integrates policy optimization with TD value estimation. 
The actor, with parameters $\theta\in\mathbb{R}^{n_\theta}$, represents a stochastic policy $\pi_\theta(a \mid s)$ and it is responsible for choosing actions. The critic, with parameters $\phi\in\mathbb{R}^{n_\phi}$, estimates a value function that evaluates the quality of the actor's actions. While the critic is typically trained using TD updates to reduce prediction error, the actor is updated via policy gradients that rely on the critic's evaluation as a performance signal. 
Let \(N \in \mathbb{N}\) denote the number of sampled trajectories and \(L \in \mathbb{N}\) the episode length. The actor parameters \(\theta\) are optimized to maximize the expected discounted return \(J(\pi_\theta)\in\mathbb{R}\) via the policy-gradient estimate
\begin{equation}
\nabla_{\theta} J(\pi_{\theta}) =
\frac{1}{N} \sum_{i=1}^{N} \sum_{t=1}^{L}
\nabla_{\theta} \log \pi_{\theta}(a_t^i \mid s_t^i)\,
A_{\phi}(s_t^i, a_t^i),
\end{equation}
where the advantage function \(A_{\phi}\) is approximated using the TD error, $A_{\phi}(s_t,a_t) \approx r_t + \gamma V_{\phi}(s_{t+1}) - V_{\phi}(s_t)$.
Here, \(s_t^i\) and \(a_t^i\) denote the state and action of the \(i\)-th sampled trajectory at time \(t\), respectively, and actions are sampled from the stochastic policy as \(a_t \sim \pi_{\theta}(\cdot \mid s_t)\).
\subsection{Actor-Critic Model Predictive Control}
\label{subsec:acmpc}
In the AC-MPC architecture~\cite{romero2024icra,Romero_2026}, the MPC module is embedded as the final layer of the actor network within a standard actor-critic pipeline trained with proximal policy optimization. By casting MPC as a differentiable layer, gradients can be propagated through the underlying optimal control problem, tightly coupling policy learning with model-based control.
Instead of manually specifying task-dependent cost functions, the MPC objective is parameterized by a neural network that outputs quadratic cost coefficients at each time step. This learned cost representation allows the control objective to be shaped directly by the reinforcement signal while preserving key advantages of MPC, including constraint handling and generalization across the state space.
To promote exploration during training, control actions are sampled around the MPC solution via a stochastic policy. At deployment time, the deterministic MPC solution is applied directly, retaining the structure of a model-based controller. Algorithm~\ref{alg:acmpc-deploy} summarizes the resulting AC-MPC workflow.

In detail, the AC-MPC framework combines two aspects: it is \emph{economic} in the objective design and \emph{differentiable} in the solver implementation. 
The cost structured adopted by the AC-MPC is~\eqref{eq:stage-cost}~\cite{romero2024icra}.
At each timestep, the actor network maps the current observation to the quadratic cost parameters \((C_t,c_t)\), defining the stage objective optimized by the MPC controller online. 
In parallel, the differentiable iLQR-based solver lets gradients of the RL loss propagate back through the MPC layer, enabling end-to-end actor optimization in an actor-critic setting. Thus, economic specifies what is optimized at runtime, while differentiable specifies how the policy is trained via the underlying optimization.

A key limitation of this architecture is computational overhead: each policy evaluation requires solving an MPC problem, and end-to-end training additionally requires differentiating through the solver---typically via the solution of an auxiliary linear system---to propagate gradients. As a result, AC-MPC incurs substantially higher training and inference latency than a standard feed-forward MLP. Although PyTorch provides highly optimized CUDA kernels for many tensor operations~\cite{amos_mpc_pytorch}, differentiable MPC workloads are often composed of repeated horizon-wise recursions and numerous small-to-medium tensor operations. In this regime, end-to-end performance is frequently limited not by raw GPU throughput, but by Python-side orchestration and kernel launch overhead associated with many short-running kernels.

\begin{algorithm}[t!]
\caption{AC--MPC deployment.}
\label{alg:acmpc-deploy}
\begin{algorithmic}[1]
\Require Trained actor $\pi_\theta$, dynamics $f$, horizon $T$, iLQR iterations $K$, bounds $\mathcal{U}$
\For{$t = 0,1,2,\ldots$}
  \State Observe state $x_t$
  \State $\{(C_t,c_t)\}_{0:T-1} \gets \pi_\theta(x_t)$
    \Comment{stage-cost parameters over the horizon}
  \State$u_t^\star \gets \textsc{DiffMPC}\!\left(x_t,\{(C_t,c_t\}_{0:T-1},f,\mathcal{U},K\right)$
    \Comment{\textsc{DiffMPC}: iLQR forward solve \cite{amos2018differentiablempc}}
  \State Apply $u_t^\star$ to the system
    \Comment{receding horizon}
\EndFor
\end{algorithmic}
\end{algorithm}


\section{Cuda-Accelerated Actor-Critic Model Predictive Control}\label{sec:prop_sol}
Motivated by the computational bottlenecks highlighted in Section~\ref{subsec:acmpc}, we propose the CA-AC-MPC, a CUDA-accelerated extension of the AC-MPC. 
Rather than relying solely on standard tensor-level dispatch, the proposed implementation leverages fused custom C++/CUDA operators that merge multiple computational steps into single kernel launches. 
Although graph-based batching (e.g., CUDA Graphs) can also mitigate these costs when tensor shapes and control flow are sufficiently static, our design targets the differentiable MPC workload directly and yields a practical reduction in end-to-end execution time for both training and inference~\cite{pytorch_cuda_graphs,pytorch_cuda_semantics}.

We consider the same box-constrained control problem adopted by the reference differentiable MPC implementation in~\cite{amos_mpc_pytorch}, which we denote hereafter as the baseline backend DiffMPC\footnote{In~\cite{amos_mpc_pytorch}, this backend is referred to as \texttt{mpc.pytorch}.}. 
The resulting control problem is
\begin{equation}\label{eq:mpc}
\begin{aligned}
\underset{x_t,u_t}{\operatorname{min}}\quad
 &\sum_{t=0}^{T-1} \frac{1}{2}z_t^\top C_t z_t + c_t^\top z_t \\
\text{s.t.}\quad
& x_{t+1} = f(x_t,u_t),\qquad x_0 = x_{\mathrm{init}}, \\
& u_{\min} \le u_t \le u_{\max},\qquad t=0,\dots,T-1,
\end{aligned}
\end{equation}
where $(C_t,c_t)$ parametrize the quadratic surrogate cost predicted by the actor network. Gradients of an external loss $\mathcal{L}$ with respect to $(C_t,c_t)$ and $x_{\mathrm{init}}$ are recovered via implicit differentiation of the KKT conditions at the converged solution~\cite{amos2018differentiablempc}, making the backward pass independent of the number of solver iterations. The forward solver follows the control-limited iLQR formulation~\cite{tassa2014control}. 
During AC-MPC training with proximal policy optimization (PPO), the forward solve is invoked at every rollout step; for a single training run with horizon \(T\) and \(K\) iLQR iterations, this results in millions of MPC solves. In DiffMPC~\cite{amos_mpc_pytorch}, each iLQR phase (linearization, Riccati backward sweep, and line-search rollout) is implemented via a Python-level \texttt{for} loop over the horizon \(T\), triggering \(\mathcal{O}(T)\) GPU kernel launches per phase and per iteration. 
When \(n_x\) and \(n_u\) are small to moderate (e.g., below \(10^2\)), each operation completes in a few microseconds, so host-side dispatch and kernel launch overhead can dominate overall runtime.
Our contribution is a C++/CUDA extension that fuses each iLQR
phase into a single kernel launch, eliminating this dispatch overhead while
retaining the same mathematical formulation and implicit differentiation targets.

The underlying solver follows a standard iLQR structure and it is implemented as a three-stage loop, described below.
These stages are mapped to three fused CUDA kernels, invoked sequentially at each iLQR iteration. 

\subsubsection{Rollout and linearization}
Given the current nominal inputs, the kernel computes the state rollout
$x_{t+1}=f(x_t,u_t)$ and the analytical Jacobians
$A_t=\partial f/\partial x \in \mathbb{R}^{n_x \times n_x}$ and $B_t=\partial f/\partial u \in \mathbb{R}^{n_x \times n_u}$
for all batch elements and timesteps in a single pass.
Analytical derivatives avoid storing large autograd graphs (i.e., the data structures employed by automatic differentiation to track operations for gradient computation), reducing memory usage and directly providing the linear time-varying model required for the iLQR backward pass.

\subsubsection{Backward pass with input bounds}
The second kernel runs the iLQR/DDP backward pass (Riccati recursion) over the horizon, yielding $\delta u_t = K_t \delta x_t + k_t$ (unconstrained) where \(K_t \in \mathbb{R}^{n_u \times n_x}\) and \(k_t \in \mathbb{R}^{n_u}\) are the feedback and feedforward gains, respectively. With box constraints $u_{\min}\le u_t\le u_{\max}$, the stagewise control update is obtained by solving the corresponding box-constrained quadratic subproblem~\cite{tassa2014control} in CUDA. The recursion is sequential in time and parallelized across batch elements.

\subsubsection{Forward pass with line-search}
We denote by $\alpha \in (0,1]$ the scalar line-search coefficient used to scale the feedforward term. In practice, the third kernel evaluates a small set of candidate values, $\alpha \in \{1.0,\,0.5,\,0.25,\,0.1\}$, in parallel. For each candidate, the kernel
applies the affine feedback update
and rolls out the dynamics to obtain the new trajectory and cost. The best
candidate is selected and used to update the nominal solution for the next iLQR
iteration.

The three stages outlined above constitute our core contribution, the CA-DiffMPC. 
Compared with a Python-level recursion over time (which typically incurs \(\mathcal{O}(T)\) GPU operator launches per phase), CA-DiffMPC requires only three kernel launches per iLQR iteration.
Embedding CA-DiffMPC as the final differentiable layer of an actor-critic network yields the overall framework, CA-AC-MPC, whose workflow is summarized in Algorithm~\ref{alg:acmpc-deploy-fast}.
The mathematical formulation and implicit differentiation strategy of the proposed approach are unchanged with respect to DiffMPC~\cite{amos2018differentiablempc,amos_mpc_pytorch}. Our contribution is thus primarily architectural: we replace per-timestep Python-level loops with three fused CUDA kernel launches per iLQR iteration.
 \begin{algorithm}[t!]
   \caption{CA-AC-MPC deployment.}
  \label{alg:acmpc-deploy-fast}
\begin{algorithmic}[1]
  \Require Actor $\pi_\theta$, dynamics $f$, horizon $T$, max iter $K$, bounds $\mathcal{U}$
  \State Initialize \textsc{CA-DiffMPC} backend
  \For{$t = 0,1,2,\ldots$}
    \State Observe state $x_t$
    \State $(C, c)_{0:T-1} \gets \pi_\theta(x_t)$ \Comment{Cost parameters}
    \State $(X^{0},U^{0}) \gets \textsc{Rollout}(x_t,f)$
    \For{$k = 1 \dots K$}
      \State $(A,B,\ell)_{0:T-1} \gets \textsc{LQ}(f, (C,c), X^{k-1}, U^{k-1})$
      \Comment{Quadraticize cost function}
      \State $(K,k)_{0:T-1} \gets \textsc{FBP}((A,B,\ell), \mathcal{U})$ \Comment{Fused backward pass}
      \State $(X^{k},U^{k}) \gets \textsc{FFP}(x_t, U^{k-1}, (K,k))$ \Comment{Fused forward pass + line-search}
      \If{\textsc{Converged}$(X^{k},U^{k})$}
        \State \textbf{break}
      \EndIf
    \EndFor
    \State $u_t^\star \gets U^{k}[0]$ \Comment{Extract current action}
    \State Apply $u_t^\star$ to system \Comment{Receding horizon}
  \EndFor
  \end{algorithmic}
  \end{algorithm}

\section{Numerical Evaluation}\label{sec:case_studies}
In this section, we numerically evaluate the proposed CA-AC-MPC framework and compare it with state-of-the-art methods. We assess task performance in a simulated drone racing scenario following~\cite{romero2022tro,kaufmann2023champion,Romero_2026}. Since the exact SplitS track geometry used in prior studies is not publicly available, we train and evaluate on a reconstructed layout based on available descriptions\footnote{Gates were reconstructed from top-down views and metric scales reported in the cited papers; gate heights are more uncertain.}. For consistency, we report lap-time comparisons only within our environment and do not compare absolute lap times with those in the literature. The reconstructed track is available in our provided code.

The comparison evaluation is thus organized considering four metrics: $(\mathbb{M}_1)$ standalone CA-DiffMPC forward- and backward-pass latencies benchmarked against DiffMPC; $(\mathbb{M}_2)$ CA-AC-MPC policy training time; $(\mathbb{M}_3)$ CA-AC-MPC policy inference time; and $(\mathbb{M}_4)$ closed-loop task performance. 
Metrics $\mathbb{M}_2\--\mathbb{M}_4$ are evaluated across three MPC horizon lengths (\(T \in \{2,5,10\}\)) and three iLQR iteration counts (\(K \in \{1,5,10\}\)). 
Metrics $\mathbb{M}_1\--\mathbb{M}_3$ are reported to quantify the computational advantages of the proposed implementation. In $\mathbb{M}_1$, we compare solver latency in isolation by benchmarking the CA-DiffMPC against the baseline DiffMPC, without integrating the actor-critic components. Metrics $\mathbb{M}_2$ and $\mathbb{M}_3$ evaluate the end-to-end training and inference latency of the complete CA-AC-MPC stack against the original AC-MPC implementation of~\cite{Romero_2026}.
However, it is worth citing that, during these comparisons, we observed persistent memory leaks when running the public AC-MPC codebase~\cite{Romero_2026} with the DiffMPC backend\footnote{See: \url{https://github.com/uzh-rpg/acmpc_public/issues/1}}. 
A practical consequence of this issue is that the baseline AC-MPC policy could not be trained to converge in our setup, preventing a direct, like-for-like, closed-loop comparison in metrics $\mathbb{M}_4$. 


Regarding the learning setup, the actor predicts the stage-cost coefficients via a neural cost map implemented as a two-layer MLP---$512$ hidden units per layer, rectified linear unit (ReLU) activations---with a sigmoid output layer that enforces the same scaling and bounds as the reference implementation. The critic is a two-layer ReLU MLP with $512$ hidden units per layer and outputs a scalar value. All other implementation details follow the original AC-MPC formulation and its public codebase. Training hyperparameters are reported in Table~\ref{tab:hyperparameters}.
All simulations were conducted on a desktop workstation equipped with an Intel Core i9-13900K CPU and an NVIDIA RTX A6000 GPU.
\begin{table}[t!]
    \centering
    \caption{Training hyperparameters values.}
    \label{tab:hyperparameters}
    \begin{tabular}{lr}
        \toprule
        \textbf{Setting} & \textbf{Value} \\
        \midrule
        Discount factor $\gamma$ & $0.99$ \\
        GAE $\lambda$ & $0.95$ \\
        Steps per update & $256$ \\
        Mini-batch size & $2048$ \\
        SGD epochs & $10$ \\
        Clip range & $0.2$ \\
        Learning rate & $3\times10^{-4} \rightarrow 3\times10^{-5}$ \\
        Entropy coef. & $0.0$ \\
        Value-loss coef. & $0.5$ \\
        Grad. clip & $0.5$ \\
        \bottomrule
    \end{tabular}
\end{table}

\subsection{Solver latency comparison ($\mathbb{M}_1$)}
To evaluate $\mathbb{M}_1$ we did not consider the drone racing scenario. 
Here, the MPC aims at stabilizing the drone at random target hovering positions, and the time required to compute each control update is recorded. The task is performed for three prediction horizons ($T \in \{2,10,50\}$), both for a single environment ($B=1$) and for a batch of $B=256$ environments to characterize scaling under parallel execution.
Table~\ref{tab:speedup} reports the corresponding numerical values. The proposed CA-DiffMPC achieves consistently low forward-pass latency in the single-instance setting compared to the baseline DiffMPC and demonstrates effective scaling when evaluated on large batches. 

\begin{table}[!t]
\centering
\caption{Execution time comparison (in milliseconds) between CA-DiffMPC and DiffMPC.}
\label{tab:speedup}
\begin{tabular}{rrcccc}
\toprule
& & \multicolumn{2}{c}{Forward time} & \multicolumn{2}{c}{Backward time} \\
\cmidrule(lr){3-4} \cmidrule(lr){5-6}
$B$ & $T$ 
& CA-DiffMPC & DiffMPC 
& CA-DiffMPC & DiffMPC \\
\midrule
\multirow{3}{*}{1}
 & 2  & $0.93$  & $9.92$  & $0.32$  & $4.81$  \\
 & 10 & $1.08$  & $41.0$  & $0.80$  & $9.16$  \\
 & 50 & $1.84$  & $492$   & $2.35$  & $30.1$  \\
\midrule
\multirow{3}{*}{256}
 & 2  & $1.21$  & $25.4$  & $0.38$  & $9.71$  \\
 & 10 & $1.68$  & $90.3$  & $0.87$  & $14.5$  \\
 & 50 & $4.00$  & $674$   & $2.60$  & $37.3$  \\
\bottomrule
\end{tabular}
\end{table}
 
\subsection{Training time comparison  ($\mathbb{M}_2$)}\label{sec:training_time_comparison}
\begin{figure}[t!]
\centering
\includegraphics[trim={0.0cm 0.0cm 0.0cm 0.0cm},clip,width=0.8\columnwidth]{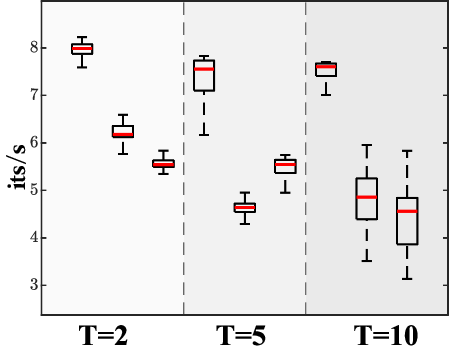}
\caption{CA-AC-MPC training throughput, measured in thousands of iterations per second. Results are grouped by prediction horizon, from left to right ($T \in \{2, 5, 10\}$). Within each group, bars represent varying numbers of iLQR iterations ($K \in \{1, 5, 10\}$, from left to right, respectively).}
\label{fig:mpciterationsspeed}
\vspace{-0.2cm}
\end{figure}

Here we performed shortened training runs under identical conditions for both solvers. In preliminary evaluations with AC-MPC (which uses DiffMPC as the differentiable MPC backend~\cite{amos_mpc_pytorch,amos2018differentiablempc}), random access memory usage increased steadily until the process terminated, preventing completion of the full \(10^{7}\) training steps. We attribute this behaviour to two likely causes: (i)~the \texttt{LQRStep} closure factory instantiated a new \texttt{torch.autograd.Function} subclass on each call, causing unbounded accumulation of class-level metadata in PyTorch's autograd engine; (ii)~the iLQR iteration loop executed with gradient tracking enabled, constructing large autograd graphs during each of the ${\sim}80$ MPC solves per PPO update that were never backpropagated through. 
We patched the library by replacing the closure factory with a single module-level \texttt{Function} class, executing the iteration loop within a \texttt{torch.no\_grad()} context, and invoking \texttt{malloc\_trim} after each rollout to reclaim fragmented heap memory. These modifications remove only unnecessary autograd bookkeeping and do not affect the solver's numerical output. 
With these patches, memory usage plateaus, and the full training completes successfully. 
The resulting training times are reported in Table~\ref{tab:training_time_matrix}.
Across both horizons, our implementation (see Fig.~\ref{fig:mpciterationsspeed} for CA-AC-MPC training throughput) consistently reduces training time relative to AC-MPC, decreasing runtime from more than two hours to about $27$ minutes (about $5.4\times$ speedup). 
For reference, training a policy based solely on an AC-MLP (without the MPC in the loop) requires roughly $16$ minutes under the same hardware setup. While this indicates that MPC-based training remains more computationally demanding, the overhead introduced by our solver is limited to a moderate increase (about $1.3\times$ with respect to the AC-MLP). 
For the sake of transparency, we note that, under our setup, the AC-MPC absolute training times differ significantly from those reported in~\cite{Romero_2026}. This discrepancy is likely due to differences in stopping criteria, hardware, and other implementation details not stated in public repositories. 
Moreover, since the AC-MPC baseline did not reliably converge in our simulations, the only result reported is the one achieving convergence, also allowing for inference time comparison.

\begin{table}[t!]
    \centering
    \caption{Policy training times for the AC-MLP, the AC-MPC, and the CA-AC-MPC.}
    \label{tab:training_time_matrix}
    \begin{tabular}{lcccc}
        \toprule
        AC-MLP & \multicolumn{4}{c}{ 16m:18s} \\
        \midrule
        \ & T & $K=1$ & $K=5$ & $K=10$ \\
        \midrule
        AC-MPC & $2$ & --      & 2h:15m:03s & -- \\
        \midrule
        CA-AC-MPC &$2$             & 21m:22s & 27m:32s & 30m:30s \\
        CA-AC-MPC &$5$             & 21m:55s & 37m:03s & 29m:01s \\
        CA-AC-MPC &$10$           & 21m:34s  & 30m:29s & 33m:22s \\
        \bottomrule
    \end{tabular}
\end{table}

\subsection{Policy Inference Times ($\mathbb{M}_3$)}
The proposed solver achieves substantial improvements in policy inference latency across all tested configurations (Table~\ref{tab:acmpcinferencetimes}). 
Relative to the inference times reported in~\cite{romero2024icra, Romero_2026}, we observe an approximately $10\times$ speedup in the baseline AC-MPC configuration ($T=2$, five iLQR iteration). For longer horizons and multiple quadratic iterations, the performance improvements become even more pronounced. 
The improved scaling stems from the constant number of GPU kernel launches in our implementation, which avoids the horizon-dependent overhead present in previous approaches. As a result, inference time grows significantly more slowly with problem complexity.
For the same reasons reported at the end of Section~\ref{sec:training_time_comparison}, comparison with AC-MPC baseline was possible only with the case achieving convergence. 

\begin{table}[t!]
    \centering
    \caption{Policy inference times (in milliseconds) for the AC-MLP, the AC-MPC, and the CA-AC-MPC.}
    \label{tab:acmpcinferencetimes}
    \begin{tabular}{lcccc}
        \toprule
        AC-MLP & \multicolumn{4}{c}{ 0.245 $\pm$ 0.01} \\
        \midrule
        \ & T & $K=1$ & $K=5$ & $K=10$ \\
        \midrule
        AC-MPC & $2$   & --         & 11.6 $\pm$ 0.9  & -- \\
        \midrule
        CA-AC-MPC & $2$                   & 1.26 $\pm$ 0.06 & 1.26 $\pm$ 0.06 & 1.26 $\pm$ 0.07 \\
        CA-AC-MPC & $5$                   & 1.43 $\pm$ 0.08 & 2.39 $\pm$ 0.08 & 2.32 $\pm$ 0.17 \\
        CA-AC-MPC & $10$                  & 1.60 $\pm$ 0.01 & 2.83 $\pm$ 0.10 & 2.99 $\pm$ 0.30 \\
        \bottomrule
    \end{tabular}
\end{table}
\subsection{Policy Performance  ($\mathbb{M}_4$)}
\begin{figure}[t!]
    \centering
    \begin{subfigure}{0.8\columnwidth}
        \centering
        \includegraphics[width=\linewidth, trim={0 15 0 0},
            clip]{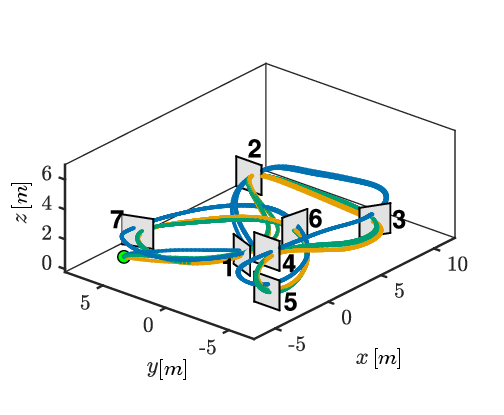}
    \end{subfigure}
    \vspace{-0.5em}
    \begin{subfigure}{0.8\columnwidth}
        \centering
        \includegraphics[width=\linewidth, trim={0 0 0 20},
            clip]{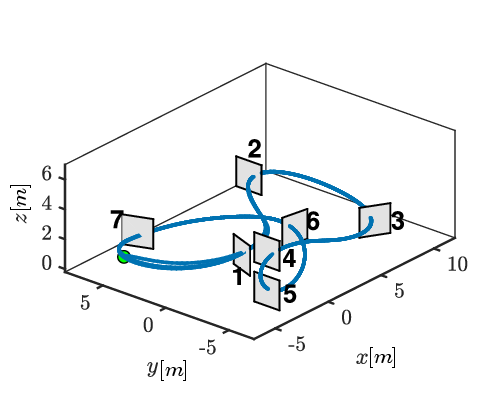}
    \end{subfigure}
    \caption{Simulated agile drone flight trajectories for horizon \(T=2\) (top) under different iLQR iteration counts (\(K\in\{1,5,10\}\), shown in blue, orange, and green, respectively) and \(T=5\) (bottom). The green circle indicates the starting position. Gates are numbered.}
    \label{fig:trajectories}
    \vspace{-0.2cm}
\end{figure}

\begin{figure*}
   \centering
\includegraphics[width=0.9\linewidth]{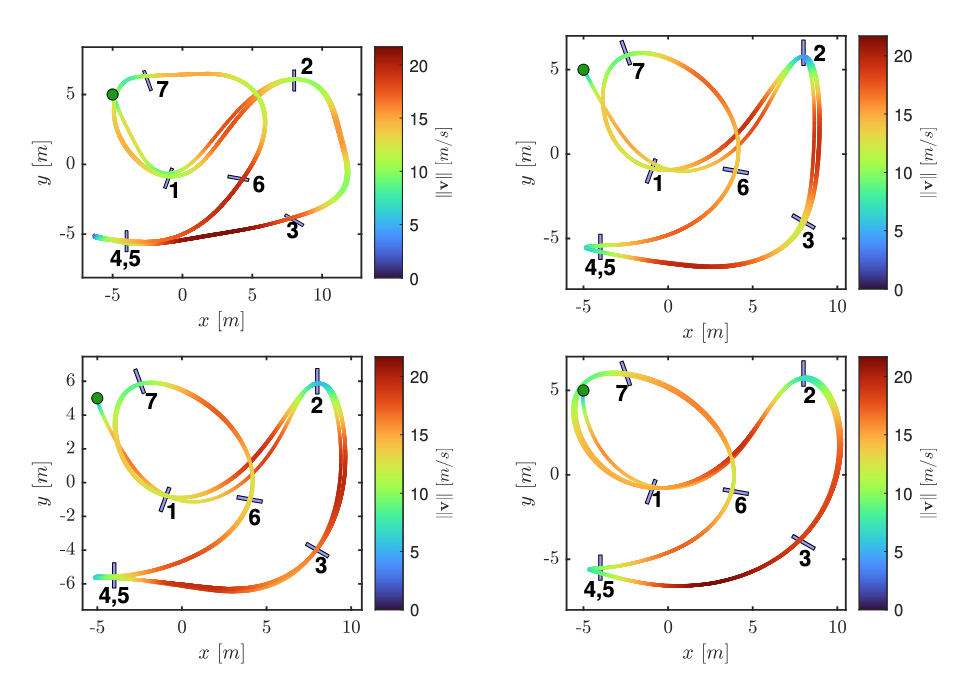}
    \caption{Top view of simulated agile drone trajectories across distinct horizons and iLQR iterations: (a) \(T=2\), \(K=1\); (b) \(T=2\), \(K=5\); (c) \(T=2\), \(K=10\); (d) \(T=5\), \(K=5\). The green circle indicates the starting position. The colorbar indicates variations in the instantaneous velocity norm. Gates are numbered.} 
\label{fig:4subfig}
\vspace{-0.5cm}
\end{figure*}

This section evaluates the performance of the trained policy across different prediction horizons. The resulting trajectories are illustrated in Fig.~\ref{fig:trajectories}, while a top-down view featuring a velocity-based color map is provided in Fig.~\ref{fig:4subfig}. Although the peak velocity remains comparable across the evaluated trajectories, the optimal policy maintains a consistently higher speed throughout the entire track. This is particularly evident between gates $2$ and $4$, before the split-S maneuver, whereas the alternative methods exhibit noticeable deceleration when approaching the waypoints.

As shown in Fig.~\ref{fig:4subfig}, all trained policies achieve peak velocities of approximately $20\,\mathrm{m/s}$, demonstrating the ability to operate the platform near its dynamic limits. The primary qualitative difference among the policies lies in the geometry of the resulting trajectories: increasing the number of iterations leads to progressively smoother paths. 

Table~\ref{tab:lap_times} reports the achieved lap times to provide a comprehensive assessment of closed-loop performance. As said, direct comparisons with~\cite{romero2024icra, Romero_2026} are intentionally omitted because the same SplitS track geometry was not publicly available. Consequently, the reported results are intended as controlled evaluation of the performance gains unlocked by the proposed solver, whose improved computational efficiency enables the use of longer MPC horizons and more solver iterations.
The most stable and effective results were obtained for horizons $T=2$ and $T=5$. 
Increasing the horizon to $T=10$, the resulting MPC problem requires the prediction of $140$ optimization variables, increasing the problem complexity. Given the limited capacity of the adopted neural network architecture, this led to training instability and ultimately prevented to a policy convergence.

\begin{table}[t!]
    \centering
    \caption{Lap times (in seconds) for the AC-MLP and the CA-AC-MPC.}
    \label{tab:lap_times}
    \begin{tabular}{lccc}
        \toprule
        \ & $T$ & $K$ & Lap time [s] \\
        \midrule
        AC-MLP     & -- & -- & 5.32 \\
        CA-AC-MPC   & 2  & 5  & 5.26 \\
        CA-AC-MPC   & 2  & 10 & 5.10 \\
        CA-AC-MPC   & 5  & 5  & 4.98 \\
        \bottomrule
    \end{tabular}
\end{table}


\section{Discussion}\label{sec:limitations}
Despite the substantial reduction in solver latency enabled by CA-AC-MPC, several limitations remain.
Increasing the MPC horizon in the CA-AC-MPC framework tends to improve performance in agile tasks (see Table~\ref{tab:lap_times}) but it also increases the number of cost parameters that the actor must predict. As the horizon grows, a simple feed-forward AC-MLP
struggles to output temporally consistent costs, leading to noisy or inconsistent objectives across timesteps. This can slow down training and reduce robustness.

A promising direction is to enforce structure in the cost parametrization: a move blocking cost could be a solution, or use temporal architectures that
explicitly model time/progress dependencies~\cite{li2023surveytransformersreinforcementlearning}.
Our solver enforces box constraints on the control inputs within the iLQR/DDP update; however, general state constraints are not directly supported by the underlying differentiable MPC layer and are currently handled through differentiable barrier/penalty terms. While this preserves differentiability, it does not provide hard constraint guarantees and may increase the learning burden, as the actor must learn both task optimality and implicit constraint satisfaction. Agile flight and racing objectives vary with track progress and maneuver phase. Thus, if the MPC cost is parameterized by a purely feed-forward network, coherent time-varying cost schedules generally require explicit progress/time inputs or a sequence model; otherwise, the policy may struggle to maintain consistent objectives over long horizons.

The three-kernel-per-iteration structure also makes CA-DiffMPC a promising candidate for embedded GPU deployment. By replacing horizon-wise PyTorch recursions with a compact, horizon-independent CUDA launch pattern, the solver reduces CPU-side dispatch overhead, a key concern for small-scale MPC workloads on platforms such as NVIDIA Jetson Orin. Although hardware validation remains future work, this architecture suggests that deployment on realistic robotic compute platforms should preserve much of the observed latency advantage.
\section{Conclusion}\label{sec:conclusion}
In this work, we extend AC-MPC with CUDA-based GPU acceleration. Our resulting CA-AC-MPC substantially reduces both training and inference time relative to the baseline implementation. The speedups come from faster differentiable MPC solves across horizon lengths and batch sizes, enabled by a fused-kernel CUDA iLQR implementation for differentiable MPC, which we call CA-DiffMPC.
Overall, CA-DiffMPC yields approximately at least one order-of-magnitude speed-ups in forward and backward solve time for single instances, and at least \(20\times\) for batched problems. As a result, the proposed CA-AC-MPC increases end-to-end training time by only about \(30\%\) relative to the AC-MLP baseline, whereas the reference AC-MPC implementation requires approximately \(30\times\) the AC-MLP training time~\cite{romero2024icra}. 
This substantially reduces the latency introduced by embedding a differentiable MPC layer within the actor--critic paradigm, which constitutes the main bottleneck of the original architecture.
Crucially, these computational advantages are obtained without compromising closed-loop performance in a drone racing scenario. Although a direct, like-for-like comparison against the full baseline AC-MPC was not possible due to issues in the public repository, our simulations show that CA-AC-MPC enables near-limit dynamic behaviour consistent with~\cite{romero2024icra}. 








\bibliographystyle{ieeetr} 
\bibliography{bibliography}

@inproceedings{amos2018differentiablempc,
  title     = {Differentiable {MPC} for End-to-End Planning and Control},
  author    = {Amos, Brandon and Rodriguez, Ivan Dario Jimenez and Sacks, Jacob and Boots, Byron and Kolter, J. Zico},
  booktitle = {Advances in Neural Information Processing Systems (NeurIPS)},
  volume    = {31},
  year      = {2018}
}

@book{borrelli2017predictive,
  title     = {Predictive Control for Linear and Hybrid Systems},
  author    = {Borrelli, Francesco and Bemporad, Alberto and Morari, Manfred},
  publisher = {Cambridge University Press},
  year      = {2017}
}

@inproceedings{tassa2014control,
  title     = {Control-Limited Differential Dynamic Programming},
  author    = {Tassa, Yuval and Mansard, Nicolas and Todorov, Emo},
  booktitle = {IEEE International Conference on Robotics and Automation (ICRA)},
  pages     = {1168--1175},
  year      = {2014}
}

@inproceedings{agrawal2019differentiable,
  title     = {Differentiable Convex Optimization Layers},
  author    = {Agrawal, Akshay and Amos, Brandon and Barratt, Shane and Boyd, Stephen and Diamond, Steven and Kolter, J. Zico},
  booktitle = {Advances in Neural Information Processing Systems (NeurIPS)},
  volume    = {32},
  year      = {2019}
}

@inproceedings{amos2017optnet,
  title     = {{OptNet}: Differentiable Optimization as a Layer in Neural Networks},
  author    = {Amos, Brandon and Kolter, J. Zico},
  booktitle = {International Conference on Machine Learning (ICML)},
  pages     = {136--145},
  year      = {2017}
}

@misc{pytorch_cuda_semantics,
  author       = {{PyTorch Contributors}},
  title        = {CUDA semantics -- PyTorch Documentation},
  howpublished = {\url{https://docs.pytorch.org/docs/stable/notes/cuda.html}},
  note         = {Accessed: 2026-02-14},
  year         = {2026}
}

@misc{pytorch_cuda_graphs,
  author       = {{PyTorch Blog}},
  title        = {Accelerating PyTorch with CUDA Graphs},
  howpublished = {\url{https://pytorch.org/blog/accelerating-pytorch-with-cuda-graphs/}},
  note         = {Accessed: 2026-02-14},
  year         = {2021}
}

@misc{amos_mpc_pytorch,
  author       = {Amos, Brandon and Jimenez, Ivan and Sacks, Jacob and Boots, Byron and Kolter, J. Zico},
  title        = {{mpc.pytorch}: A fast and differentiable model predictive control solver for PyTorch (v0.0.6)},
  howpublished = {\url{https://github.com/locuslab/mpc.pytorch/tree/v0.0.6}},
  note         = {GitHub repository, MIT License. Accessed: 2026-02-14},
  year         = {2024}
}

@book{Findeisen2017EMPC,
  title     = {Economic Model Predictive Control},
  editor    = {Findeisen, Rolf and Gr{\"u}ne, Lars and M{\"u}ller, Matthias A.},
  publisher = {Springer},
  year      = {2017},
  address   = {Cham},
  isbn      = {978-3-319-55866-2},
  doi       = {10.1007/978-3-319-55867-9}
}

@ARTICLE{romero2022tro,
  author={Romero, Angel and Sun, Sihao and Foehn, Philipp and Scaramuzza, Davide},
  journal={IEEE Transactions on Robotics}, 
  title={Model Predictive Contouring Control for Time-Optimal Quadrotor Flight}, 
  year={2022},
  volume={38},
  number={6},
  pages={3340-3356},
  doi={10.1109/TRO.2022.3173711}}

@article{song2023reaching,
  title={Reaching the limit in autonomous racing: Optimal control versus reinforcement learning},
  author={Song, Yunlong and Romero, Angel and M{\"u}ller, Matthias and Koltun, Vladlen and Scaramuzza, Davide},
  journal={Science Robotics},
  volume={8},
  number={82},
  pages={eadg1462},
  year={2023},
  publisher={American Association for the Advancement of Science}
}

@article{kaufmann2023champion,
  title={Champion-level drone racing using deep reinforcement learning},
  author={Kaufmann, Elia and Bauersfeld, Leonard and Loquercio, Antonio and M{\"u}ller, Matthias and Koltun, Vladlen and Scaramuzza, Davide},
  journal={Nature},
  volume={620},
  number={7976},
  pages={982--987},
  year={2023},
  publisher={Nature Publishing Group UK London}
}

@article{Romero_2026,
   title={Actor–Critic Model Predictive Control: Differentiable Optimization Meets Reinforcement Learning for Agile Flight},
   volume={42},
   journal={IEEE Transactions on Robotics},
   publisher={Institute of Electrical and Electronics Engineers (IEEE)},
   author={Romero, Angel and Aljalbout, Elie and Song, Yunlong and Scaramuzza, Davide},
   year={2026},
   pages={673–692} }

@ARTICLE{712192,
  author={Sutton, R.S. and Barto, A.G.},
  journal={IEEE Transactions on Neural Networks}, 
  title={Reinforcement Learning: An Introduction}, 
  year={1998},
  volume={9},
  number={5},
  pages={1054-1054},
  doi={10.1109/TNN.1998.712192}}

@INPROCEEDINGS{romero2024icra,
  author={Romero, Angel and Song, Yunlong and Scaramuzza, Davide},
  booktitle={2024 IEEE International Conference on Robotics and Automation (ICRA)}, 
  title={Actor-Critic Model Predictive Control}, 
  year={2024},
  volume={},
  number={},
  pages={14777-14784},
  doi={10.1109/ICRA57147.2024.10610381}}

@article{levine2016gps,
  title={End-to-End Training of Deep Visuomotor Policies},
  author={Levine, Sergey and Finn, Chelsea and Darrell, Trevor and Abbeel, Pieter},
  journal={Journal of Machine Learning Research},
  volume={17},
  number={39},
  pages={1--40},
  year={2016}
}

@inproceedings{nagabandi2018mpc,
  title={Neural Network Dynamics for Model-Based Deep Reinforcement Learning with Model-Free Fine-Tuning},
  author={Nagabandi, Anusha and Kahn, Gregory and Fearing, Ronald S. and Levine, Sergey},
  booktitle={IEEE International Conference on Robotics and Automation (ICRA)},
  year={2018}
}

@ARTICLE{amatucciral2026,
  author={Amatucci, Lorenzo and Sousa-Pinto, João and Turrisi, Giulio and Orban, Dominique and Barasuol, Victor and Semini, Claudio},
  journal={IEEE Robotics and Automation Letters}, 
  title={Primal-Dual iLQR for GPU-Accelerated Learning and Control in Legged Robots}, 
  year={2026},
  volume={11},
  number={1},
  pages={1010-1017},
  doi={10.1109/LRA.2025.3632610}}

@article{adabag2025differentiable,
  title={Differentiable Model Predictive Control on the GPU},
  author={Adabag, Emre and Greiff, Marcus and Subosits, John and Lew, Thomas},
  journal={arXiv preprint arXiv:2510.06179},
  year={2025}
}

@book{grne2013nonlinear,
  title={Nonlinear model predictive control: theory and algorithms},
  author={Grne, Lars and Pannek, Jrgen},
  year={2013},
  publisher={Springer Publishing Company, Incorporated}
}

@book{ellis2017empc,
    author    = {Matthew Ellis and Jinfeng Liu and Panagiotis D. Christofides},
    title     = {Economic Model Predictive Control: Theory, Formulations and
  Chemical Process Applications},
    series    = {Advances in Industrial Control},
    publisher = {Springer},
    year      = {2017},
  }

@article{xie2024three,
  title={Three-dimensional variable center of mass height biped walking using a new model and nonlinear model predictive control},
  author={Xie, Zhongqu and Wang, Yulin and Luo, Xiang and Arpenti, Pierluigi and Ruggiero, Fabio and Siciliano, Bruno},
  journal={Mechanism and Machine Theory},
  volume={197},
  pages={105651},
  year={2024},
  publisher={Elsevier}
}

@misc{li2023surveytransformersreinforcementlearning,
      title={A Survey on Transformers in Reinforcement Learning}, 
      author={Wenzhe Li and Hao Luo and Zichuan Lin and Chongjie Zhang and Zongqing Lu and Deheng Ye},
      year={2023},
      eprint={2301.03044},
      archivePrefix={arXiv},
      primaryClass={cs.LG},
      url={https://arxiv.org/abs/2301.03044}, 
}
\end{document}